%% file: main.tex
\documentclass[10pt,twocolumn,letterpaper]{article}

\usepackage[]{cvpr}      
\usepackage[numbers]{natbib}


\definecolor{cvprblue}{rgb}{0.21,0.49,0.74}
\usepackage[breaklinks,colorlinks,citecolor=cvprblue]{hyperref}

\title{
DeOcc-1-to-3: 3D De-Occlusion from a Single Image via Self-Supervised Multi-View Diffusion
}

\author{
Yansong Qu\textsuperscript{1},
Shaohui Dai\textsuperscript{1},
Xinyang Li\textsuperscript{1},
Yuze Wang\textsuperscript{2},\\
You Shen\textsuperscript{1},
Liujuan Cao\textsuperscript{1},
Rongrong Ji\textsuperscript{1}\\[0.5em]
\textsuperscript{1}Key Laboratory of Multimedia Trusted Perception and Efficient Computing,\\
Ministry of Education of China, Xiamen University,\\
\textsuperscript{2}State Key Laboratory of Virtual Reality Technology and Systems, Beihang University
}

\begin{document}
\twocolumn[{
\maketitle
\begin{center}
    \includegraphics[width=0.99\linewidth]{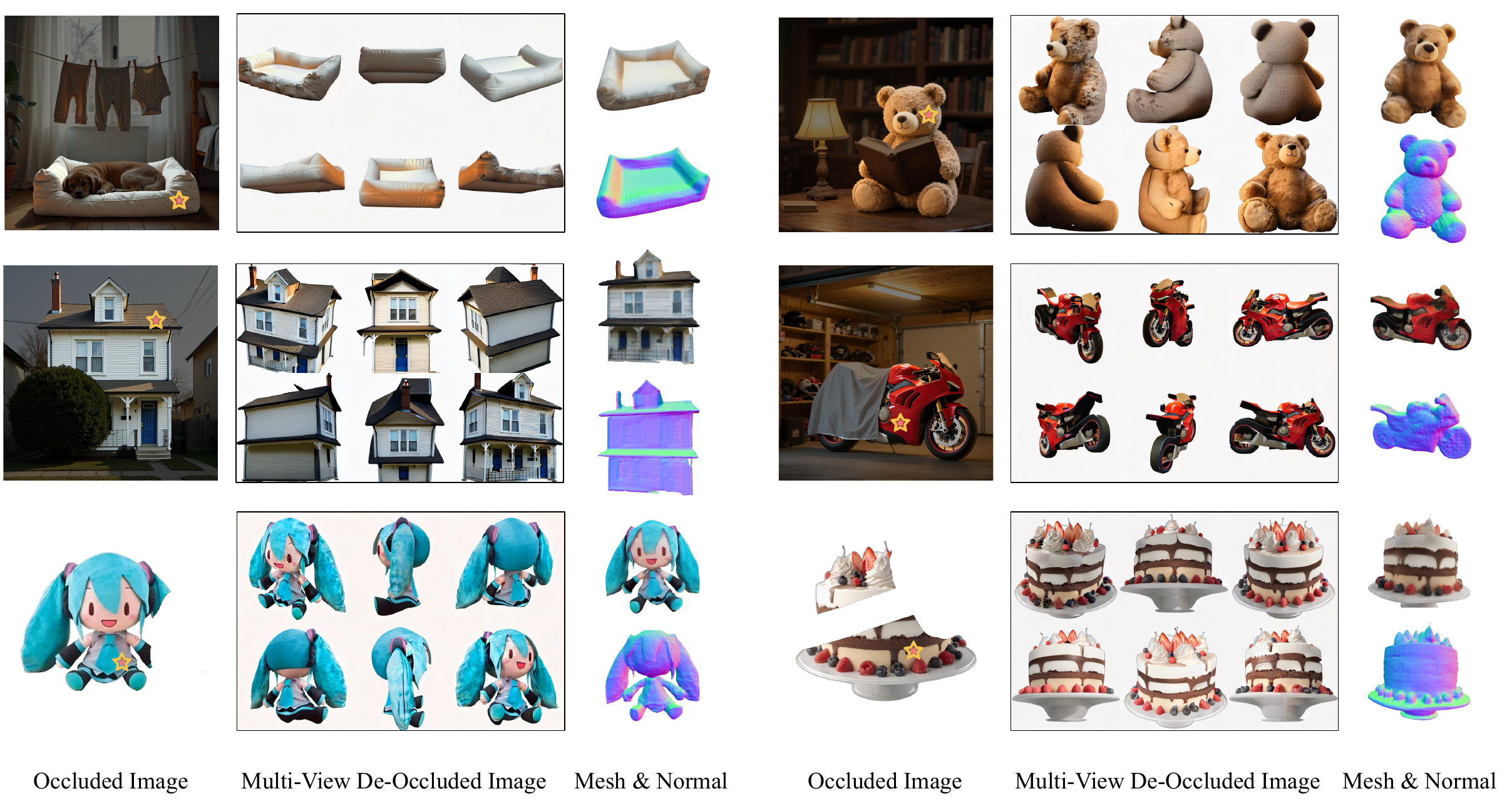}
    \label{fig:teaser}
    \captionof{figure}{\textbf{DeOcc-1-to-3} takes a single occluded image (left) as input and synthesizes structurally consistent multi-view de-occluded images (middle). These outputs can be seamlessly integrated into various 3D reconstruction or generation frameworks to produce accurate meshes and surface normals (right). The proposed pipeline demonstrates generalization across diverse object categories and occlusion scenarios.}
    \vspace{0.1cm}
\end{center}
}]

\input{sec/0_abstract}    
\input{sec/1_intro}

\input{sec/2_relatedworks}
\input{sec/3_method}

\input{sec/4_exp}
\input{sec/5_conclusion}

\small
\bibliographystyle{ieeetr}
\bibliography{main}

\end{document}

%% file: sec/0_abstract.tex
\begin{abstract}

    Reconstructing 3D objects from a single image remains challenging, especially under real-world occlusions. While recent diffusion-based view synthesis models can generate consistent novel views from a single RGB image, they typically assume fully visible inputs and fail when parts of the object are occluded, resulting in degraded 3D reconstruction quality.
    We propose DeOcc-1-to-3, an end-to-end framework for occlusion-aware multi-view generation that synthesizes six structurally consistent novel views directly from a single occluded image, enabling reliable 3D reconstruction without prior inpainting or manual annotations. Our self-supervised training pipeline leverages occluded–unoccluded image pairs and pseudo-ground-truth views to teach the model structure-aware completion and view consistency. Without modifying the original architecture, we fully fine-tune the view synthesis model to jointly learn completion and multi-view generation.
    Additionally, we introduce the first benchmark for occlusion-aware reconstruction, covering diverse occlusion levels, object categories, and masking patterns, providing a standardized protocol for future evaluation. Project page: \url{https://quyans.github.io/DeOcc123/}

\end{abstract}

%% file: sec/1_intro.tex
\section{Introduction}
\label{sec:intro}

Reconstructing a complete 3D object from a single image is a long-standing challenge in computer vision. Recent advances in diffusion-based multi-view generation models \cite{liu2023zero123,shi2023zero123pp,xu2024instantmesh,shi2023mvdream} have enabled the generation of structurally consistent novel views from a single RGB image, providing high-quality inputs for downstream 3D reconstruction. However, these models typically assume that the input object is fully visible. In practice, images often suffer from partial occlusions caused by clutter, object interactions, or limited viewpoints, making this assumption unrealistic. 
Occlusions pose a major challenge to both view synthesis and 3D modeling. When part of the object is invisible, existing methods often fail to correctly infer its geometry and appearance, resulting in inconsistent novel views 
and broken or incomplete reconstructions.

A common workaround is a two-stage pipeline: apply 2D inpainting \cite{ozguroglu2024pix2gestalt,xu2024amodal, dogaru2024generalizable, zhan2020self, zhan2024amodal,li2025synergyamodal} to complete the occluded regions, then apply view synthesis or 3D reconstruction to the completed image.  However, this approach suffers from three major limitations: (1) 2D inpainting lacks 3D priors and fails to ensure multi-view consistency; 
(2) view synthesis is unaware of uncertainties in hallucinated content, leading to artifacts; 
(3) the decoupled design prevents joint optimization, causing error accumulation across stages.

To address these challenges, we propose an end-to-end, occlusion-aware multi-view generation framework that directly predicts six structurally consistent novel views from a single occluded image. Without altering the original view synthesis architecture, we fully fine-tune the model to jointly learn occlusion completion and novel view synthesis within a unified generation process.
To eliminate the need for manual annotations, we introduce a self-supervised training strategy. We construct paired occluded and unoccluded images by randomly applying occlusions to 2D inputs. For each unoccluded image, a pretrained multi-view generation model generates six-view pseudo-ground-truths, which supervise the network when only the occluded image is available. This enables the model to learn structure-aware completion and view-consistent synthesis.

In addition, we construct a comprehensive benchmark for occlusion-aware reconstruction, spanning multiple occlusion levels, object categories, and masking patterns. This benchmark provides a standardized evaluation protocol for quantitatively assessing the performance of occlusion-aware view synthesis and 3D reconstruction methods.
In summary, our main contributions are as follows:
\begin{itemize}
\item We introduce the task of occlusion-aware multi-view generation and propose an end-to-end framework that synthesizes structurally consistent novel views directly from a single partially occluded image.
\item We develop a self-supervised training paradigm that leverages paired occluded-unoccluded images, enabling structure-aware learning without the need for manual annotations.
\item We fine-tune the multi-view generation model without modifying its architecture, achieving strong performance and seamless compatibility with downstream 3D reconstruction frameworks such as InstantMesh \cite{xu2024instantmesh}.
\item We establish the first benchmark for occlusion-aware reconstruction, encompassing diverse occlusion patterns and object categories, along with standardized evaluation metrics.
\end{itemize}

%% file: sec/2_relatedworks.tex
\section{Related Works}
\label{sec_related_works}

\subsection{Single-image to 3D Representations}

Early approaches to single-image 3D reconstruction focused on predicting explicit or implicit representations, such as meshes, point clouds, or signed distance fields (SDFs)~\cite{choy20163d,newcombe2011kinectfusion,park2019deepsdf}. However, these methods often struggle with complex geometries and occlusions. With the advent of neural rendering, NeRF~\cite{mildenhall2021nerf,wang2023rip,qu2023sg_nerf,huang2024nerf} achieved high-fidelity 3D reconstruction via differentiable volumetric rendering, and PixelNeRF~\cite{yu2021pixelnerf} extended this to single-image inputs. To improve efficiency, recent methods such as 3D Gaussian Splatting (3DGS)~\cite{kerbl20233dgs,qu2024goi,shen2025evolving,wang2025look,guo2025wildseg3d,dai2025training} adopt explicit Gaussian primitives for real-time rendering and editing. Nonetheless, regressing NeRF or 3DGS representations directly from a single view remains challenging due to geometric ambiguity and a lack of multi-view consistency.

With the rise of powerful image-conditioned diffusion models, DreamFusion~\cite{poole2022dreamfusion} and its successors~\cite{li2024director3d,qu2025drag,wang2023prolificdreamer} introduced Score Distillation Sampling (SDS), which leverages a pretrained 2D text-to-image diffusion model as an implicit energy function to guide the optimization of 3D representations. This paradigm enables text-driven 3D generation without explicit 3D supervision. While effective, SDS-based methods often suffer from high computational cost, multi-face inconsistencies, and saturated appearances. 
An alternative line of work directly fine-tunes diffusion models to synthesize structurally consistent multi-view images from a single input, as demonstrated by Zero-1-to-3~\cite{liu2023zero123}, Zero123++~\cite{shi2023zero123pp}, and MVDream~\cite{shi2023mvdream}. These approaches support efficient and high-quality 3D reconstruction via standard mesh-based pipelines. However, they assume fully visible inputs and struggle when the target object is partially occluded. 
This highlights a key limitation: existing models lack the capability to reason about and complete occluded structures, which is critical for real-world applications where occlusions are common.

\subsection{2D Amodal Completion}
2D amodal completion aims to recover the complete shape and appearance of partially occluded objects in images~\cite{li2025synergyamodal}. 
Early approaches~\cite{kimia2003euler,silberman2014contour} rely on geometric heuristics—such as Euler spirals and Bézier curves—to extrapolate occluded boundaries based on predefined occlusion orders. 
However, these methods are limited to simple shapes and lack robustness for complex real-world scenarios. 
Subsequent works~\cite{zhang2022face,yan2019visualizing,zhou2021human,papadopoulos2019make} adopt supervised learning on synthetic datasets, 
but are typically constrained to specific object categories and occlusion patterns. 
More recently, with the advancement of generative models, several methods~\cite{zhan2024amodal,duan2025diffusion,ozguroglu2024pix2gestalt,lee2025tuning} 
have addressed amodal completion using powerful image generation frameworks---such as Pix2Gestalt~\cite{ozguroglu2024pix2gestalt} and SynergyAmodal~\cite{li2025synergyamodal}---achieving promising zero-shot performance.

However, directly applying these methods to 3D reconstruction presents several limitations. First, 2D completion models are inherently limited by their lack of multi-view reasoning and may produce geometrically inconsistent outputs when extended to 3D. Second, these pipelines decouple the completion and 3D reconstruction stages, preventing joint optimization of structural consistency. 
To overcome these limitations, we propose an end-to-end framework that integrates 3D awareness into the generative process, enabling structurally consistent multi-view completions from a single occluded image.

\subsection{3D De-occlusion}

Early attempts at 3D de-occlusion often relied on 2.5D depth completion~\cite{zhang2018deep, ma2018sparse} 
or template-based human mesh fitting~\cite{alldieck2018video, pavlakos2019expressive}, which typically require RGB-D inputs, 
body priors, or multi-frame observations. While effective in constrained scenarios, these methods struggle 
to generalize to diverse object categories and complex occlusion patterns. 
Recent methods have explored generative strategies. CHROME~\cite{dutta2025chrome} employs pose-conditioned diffusion models 
to synthesize multi-view images of occluded humans, which are then reconstructed via Gaussian splatting. OccFusion~\cite{sun2024occfusion} renders coarse human meshes and refines them using diffusion-based image inpainting. Slice3D~\cite{wang2024slice3d} predicts cross-sectional slices from occluded views and assembles them into 3D volumes.

Despite promising results, these approaches often rely on dense supervision, task-specific priors, or customized architectures, limiting their scalability. In contrast, our method directly generates structure-consistent novel views from a single occluded image, and integrates seamlessly with off-the-shelf mesh-based reconstruction pipelines. 

%% file: sec/3_method.tex
\section{Method}

\begin{figure}
    \centering
    \includegraphics[width=\linewidth]{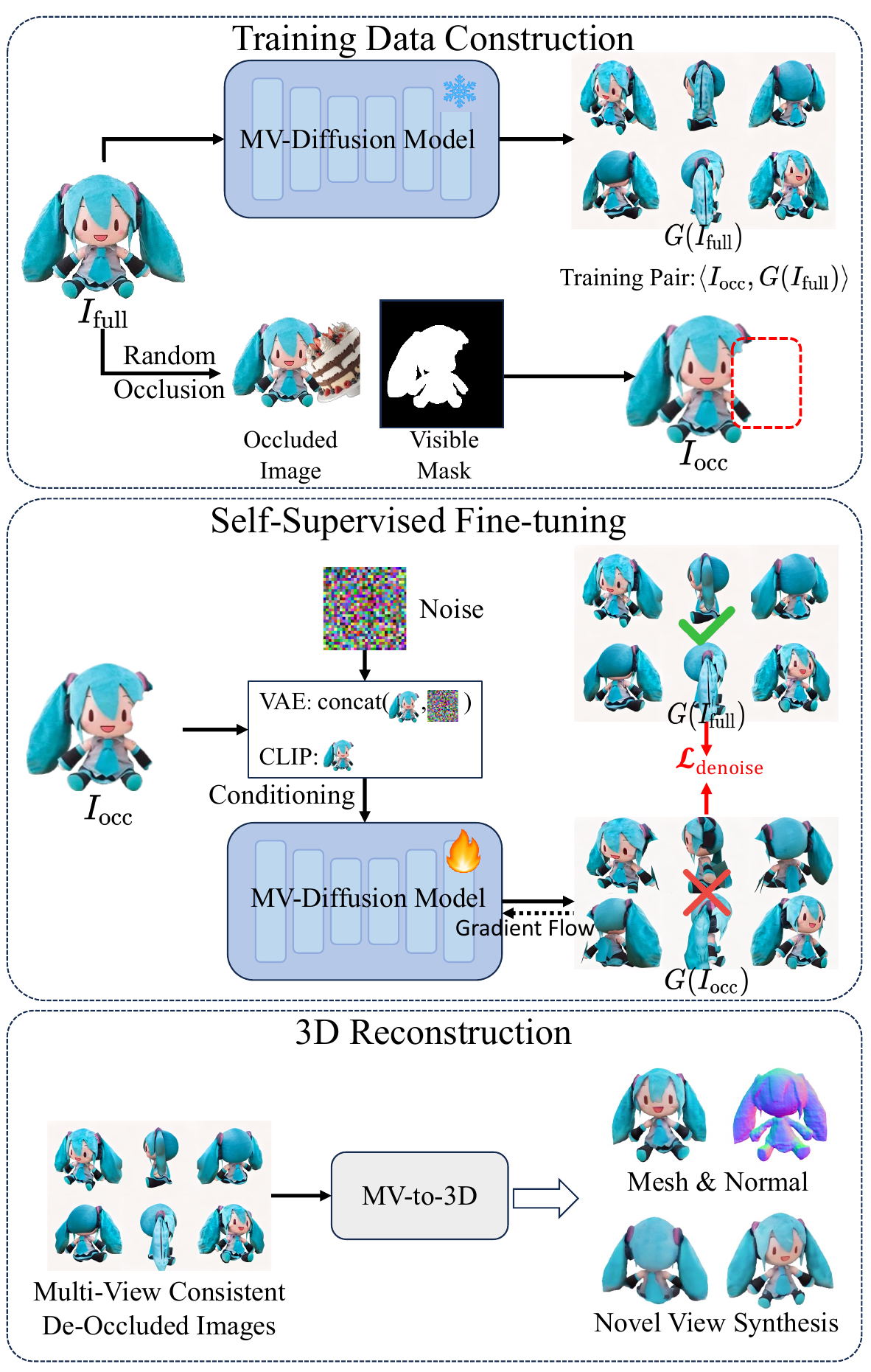}
    \caption{
    Overview of DeOcc-1-to-3.  
\textbf{Top}: Occluded images \( I_{\text{occ}} \) are generated by applying random occlusions to full images \( I_{\text{full}} \). 
A frozen multi-view diffusion model produces six-view pseudo-ground-truths \( G(I_{\text{full}}) \), forming training pairs \( \langle I_{\text{occ}}, G(I_{\text{full}}) \rangle \).  
\textbf{Middle}: The student model is fully fine-tuned to predict consistent novel views \( G(I_{\text{occ}}) \), supervised by a denoising loss \( \mathcal{L}_{\text{denoise}} \).  
\textbf{Bottom}: The predicted six-view images are fed into downstream reconstruction models for 3D reconstruction.
    }
    \label{fig:pipeline}
\end{figure}

This section presents the overall framework of our occlusion-aware multi-view generation approach, covering training data construction (Sec. \ref{sec:training-data-construction}), self-supervised fine-tuning (Sec. \ref{sec:model-structure-and-training-strategy}), and integration with downstream 3D reconstruction pipelines (Sec. \ref{sec:3d-reconstruction-integration}). Without modifying the original view synthesis architecture, our method enables structural completion and consistent novel view generation through full model fine-tuning. 
We also construct a benchmark dataset for 3D de-occlusion reconstruction to evaluate the performance of 3D de-occlusion reconstruction methods (Sec. \ref{subsec:occ-lvis-benchmark}).

\subsection{Preliminaries}

We build upon Zero123++~\cite{shi2023zero123pp}, a diffusion-based multi-view generation model that synthesizes six novel views from a single RGB image. The output is formatted as a $3 \times 2$ tiled image corresponding to six predefined camera poses, defined by:
\begin{itemize}
    \item Elevation angles: $\{30^\circ, -20^\circ\}$
    \item Azimuth angles: $\{30^\circ, 90^\circ, 150^\circ, 210^\circ, 270^\circ, 330^\circ\}$
\end{itemize}
Given an input image $I_{\text{input}}$, the model is trained with a velocity-based denoising objective:
\begin{equation}
\mathcal{L}_{\text{denoise}} = \mathbb{E}_{x_0, \epsilon, t} \left[ \| \epsilon - \epsilon_\theta(x_t, t \mid I_{\text{input}}) \|^2 \right],
\end{equation}
where $x_t$ is the noisy latent, and $\epsilon_\theta$ denotes the noise predicted by the model.

Zero123++ leverages both local conditioning (via scaled reference attention) and global conditioning (via CLIP image embeddings), enabling strong spatial coherence and semantic consistency across synthesized views. These properties make it a strong and architecture-agnostic backbone for our occlusion-aware fine-tuning framework. Moreover, as Zero123++ is trained with explicit multi-view supervision, it inherently learns multi-view consistency, making it well-suited for view-consistent generation tasks under occlusion.

\subsection{Overall Framework}
\label{sec:overall-framework}
Previous works focused on image completion from a single image followed by 3D generation, 
which often resulted in poor 3D reconstruction integrity. This is because such 2D de-occlusion 
models only consider 2D image completion without accounting for 3D structural integrity. 
Inspired by MVDream~\cite{shi2023mvdream} and Zero123++\cite{shi2023zero123pp}, we propose a native 3D de-occlusion framework based on a multi-view diffusion model. Our method constructs occlusion-augmented training pairs with pseudo multi-view supervision, and fine-tunes the view synthesis model to produce occlusion-aware novel views. The generated images are then passed to a 3D reconstruction module (e.g., InstantMesh\cite{xu2024instantmesh}) to recover a complete 3D object.
The entire pipeline is annotation-free, category-agnostic, and robust to diverse occlusion patterns. Specifically, our goal is to synthesize six predefined RGB views from a single occluded image, ensuring both structural and appearance consistency across views for effective downstream 3D reconstruction.

As illustrated in Figure~\ref{fig:pipeline}, the proposed pipeline consists of three stages:
 (1) Construct occlusion-augmented training pairs with pseudo multi-view supervision;
 (2) Fine-tune a multi-view diffusion model to enable occlusion-aware generation;
 (3) Feed the generated views into a 3D reconstruction module to obtain a complete 3D model.

\subsection{Training Data Construction}
\label{sec:training-data-construction}

The limited availability of real-world 3D data also makes it challenging to obtain universally consistent occlusion examples. While synthetic datasets~\cite{hu2019sail} can be used to generate such data, they often introduce a significant domain gap, limiting the generalization of models to real-world scenarios. In contrast, real-world 2D data is abundant and diverse.
To leverage this, we adopt a two-stage strategy: first, we construct a large-scale 2D occlusion dataset, and then use a pre-trained multi-view diffusion model to generate 3D-consistent pseudo-ground-truth views for supervision.

\textbf{2D Occlusion Data Construction.}
We construct occlusion-aware image pairs based on the SA-1B dataset~\cite{kirillov2023segment}, utilizing the Segment Anything Model (SAM)~\cite{kirillov2023segment} to segment foreground objects. A randomly selected segmented object is overlaid onto a natural background to synthesize occlusion scenarios. This process produces:
\begin{itemize}
    \item a raw image containing the complete foreground object, \( I_{\text{raw}} \),
    \item the full foreground object mask, \( M_{\text{full}} \),
    \item an occluded composite image, \( I_{\text{mix}} \),
    \item and the mask of the visible (unoccluded) part of the target object, \( M_{\text{occ}} \).
\end{itemize}
We then compute the paired foreground images as:
\begin{align*}
    I_{\text{full}} &= I_{\text{raw}} \odot M_{\text{full}}, \\
    I_{\text{occ}} &= I_{\text{mix}} \odot M_{\text{occ}},
\end{align*}
where \(\odot\) denotes element-wise multiplication.

We also apply filtering to remove occlusion samples whose foreground objects are intrinsically 
incomplete, preventing the model from learning to reproduce incomplete shapes \cite{ozguroglu2024pix2gestalt}. 
Furthermore, because our 2D dataset are used for predicting pseudo-ground truth for the subsequent 
3D reconstruction, we curate it to match the input distribution of the employed 
multi-view diffusion model. In particular, we exclude samples where a complete 
foreground object lies at the image boundary. The final dataset contains 100K samples.

\textbf{3D Occlusion Data Construction.}
We first input the clean image \(I_{\text{full}}\) into a pretrained multi-view diffusion model \(G\) (used as a teacher) 
to generate six-view pseudo-ground-truth images \(G(I_{\text{full}})\) with high structural consistency. 
This enables the construction of paired training samples \(\langle I_{\text{occ}}, G(I_{\text{full}}) \rangle\) for 3D de-occlusion learning.
Then, we use \(I_{\text{occ}}\) as the conditional input to fine-tune the same model (student) to synthesize the corresponding novel views directly.

\begin{figure*}[h]
    \centering
    \includegraphics[width=\linewidth]{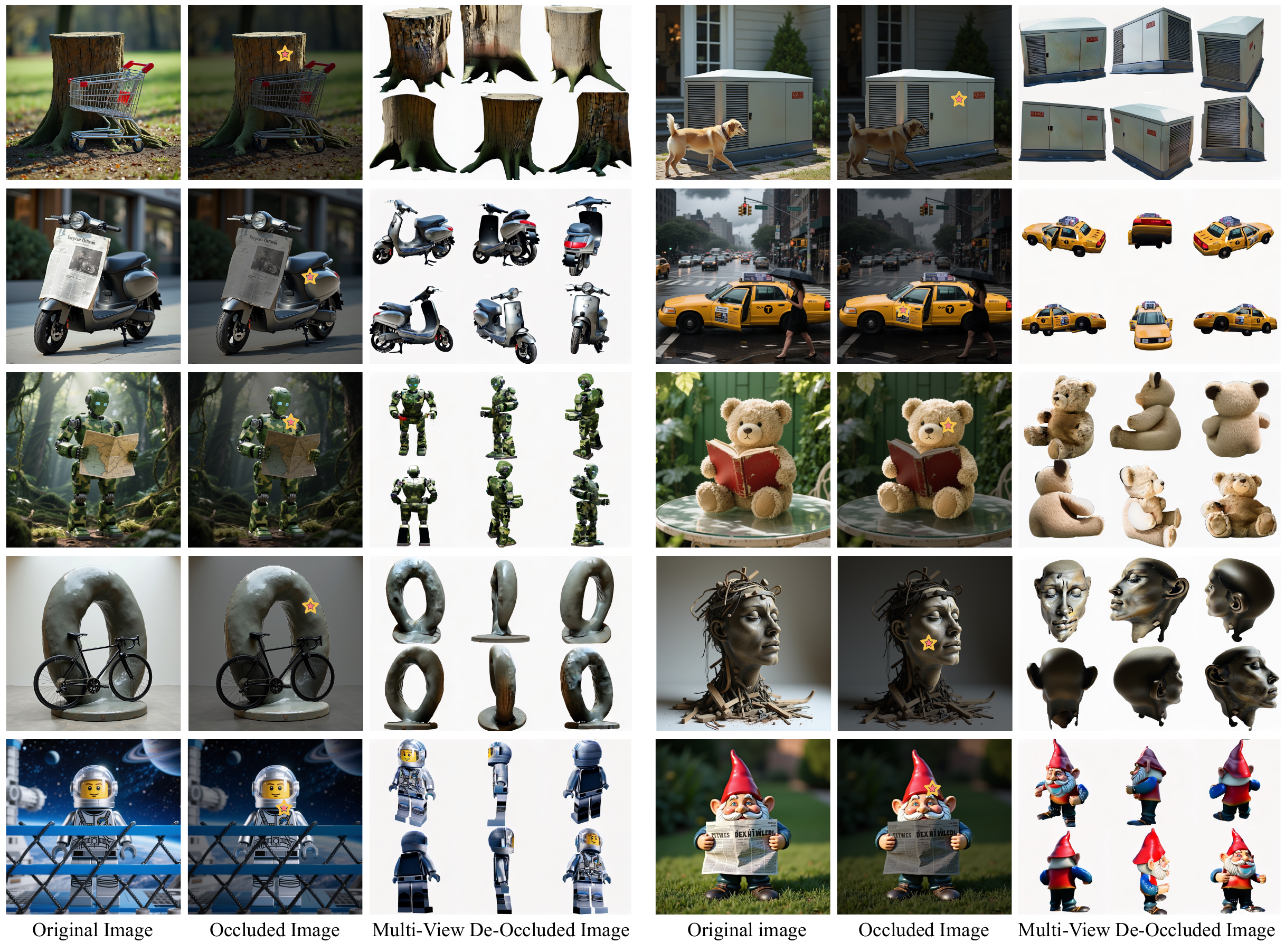}
    \caption{
Qualitative de-occlusion results on diverse objects. 
Each triplet shows (left) the original image, (middle) the occluded input, and (right) our multi-view de-occluded output (six views). 
Our method recovers coherent geometry and texture across various shapes, materials, and occlusion types.
}	
    \label{fig:results}
\end{figure*}

To enhance robustness and prevent overfitting to any specific occlusion patterns, we further augment the training set in two ways:
(1) we apply both dilation and erosion to the occlusion masks to simulate a broader range of occlusion severities and shapes, 
enabling the model to learn from both heavier and lighter occlusion scenarios; these augmented samples are included in the training set to improve robustness;
(2) we also include a subset of unoccluded foreground objects \(\langle I_{\text{full}}, G(I_{\text{full}}) \rangle\) 
as identity pairs in the training data to prevent the model from hallucinating unnecessary completions when the input is already complete.

After filtering out defective or ambiguous samples, the final 3D de-occlusion dataset contains approximately 40K high-quality training pairs, 
spanning a wide range of object categories, occlusion levels, and masking patterns. This pipeline is fully self-supervised 
and requires no manual labels such as segmentation masks or depth maps.

\subsection{Model Structure and Training Strategy}
\label{sec:model-structure-and-training-strategy}
As illustrated in Figure~\ref{fig:pipeline}, we adopt a diffusion-based multi-view generation backbone \(G\) (e.g., Zero123++) without modifying its architecture. The model takes a single occluded RGB image \(I_{\text{occ}}\) as input and produces a $3 \times 2$ concatenated image representing six predefined novel views, denoted as \(G(I_{\text{occ}})\). 
Importantly, no text prompts are used during occlusion-aware generation, following our design principle of ensuring strong usability and generalization across categories and occlusion types.

Training is performed with a standard denoising objective:
\[
\mathcal{L}_{\text{denoise}} = \mathbb{E}_{x_0, \epsilon, t} \left[ \| \epsilon - \epsilon_\theta(x_t, t \mid I_{\text{occ}}) \|^2 \right],
\]
where \(x_t\) is a noisy version of the pseudo ground-truth \(G(I_{\text{full}})\), and \(\epsilon_\theta\) 
is the predicted noise at timestep \(t\).

We fine-tune the entire U-Net, including residual blocks and attention modules. Training uses the AdamW 
optimizer with an initial learning rate of \(2 \times 10^{-5}\), a batch size of 32, and a total of 150k steps. 
We retain the same noise schedule and both local (reference attention) and global (CLIP-based image encoder) conditions
 as in the original model to ensure training stability and generation consistency.

To further improve the robustness and visual fidelity of the generated views, we maintain an Exponential Moving Average (EMA) of the model weights \(\theta\) during training:
\[
\theta_{\text{EMA}} \leftarrow \beta \cdot \theta_{\text{EMA}} + (1 - \beta) \cdot \theta,
\]
with decay rate \(\beta = 0.9999\). The EMA weights are used at inference time to improve generation stability and quality.

\subsection{3D Reconstruction Integration}
\label{sec:3d-reconstruction-integration}

The proposed occlusion-aware multi-view generation framework synthesizes six-view RGB images that are both geometrically consistent and structurally complete. These images can serve as universal inputs for a variety of 3D reconstruction or generation methods~\cite{xu2024instantmesh,xiang2025trellis3d,zhao2025hunyuan3d}, enabling the recovery of 3D models in diverse representations, such as NeRF~\cite{mildenhall2021nerf}, 3D Gaussian Splatting~\cite{kerbl20233dgs}, and mesh-based surfaces.

In our implementation, we adopt InstantMesh~\cite{xu2024instantmesh} as a representative backend due to its high efficiency and ability to generate watertight meshes. InstantMesh extracts tri-plane features from the input views and performs sparse-view geometry reasoning to decode a 3D mesh. 
Notably, since our model maintains the same output format as the original multi-view diffusion architecture, it can be seamlessly integrated into existing pipelines without requiring any architectural modifications. Experimental results demonstrate that our occlusion-aware generator significantly enhances 3D reconstruction performance in occluded scenarios, yielding more complete, consistent, and smooth surfaces.

\subsection{Occ-LVIS Benchmark}
\label{subsec:occ-lvis-benchmark}
To systematically evaluate 3D object reconstruction under occlusions, we construct a novel benchmark 
based on the Objaverse-LVIS dataset~\cite{deitke2023objaverse}. Specifically, we utilize high-quality 
assets from the LVIS subset and render them into multi-view image sequences under controlled occlusion settings.

Each object is first rendered from a canonical frontal view, and then occluded using randomly selected foreground objects.
We then generate:
\begin{itemize}
    \item \textbf{Six canonical views} with azimuth angles of \{30°, 90°, 150°, 210°, 270°, 330°\} and alternating elevation angles of \{30°, -20°\};
    \item \textbf{Four random views} to evaluate generalization to unseen viewpoints;
    \item \textbf{Geometry-complete mesh ground truths}, for evaluating 3D reconstruction fidelity.
\end{itemize}

To facilitate analysis under varying difficulty, we further stratify the benchmark into five occlusion levels 
based on the occlusion ratio (i.e., the proportion of the target object’s visible area):
\begin{table}[h]
    \centering
    \caption{Occlusion level statistics in the \textbf{Occ-LVIS} benchmark.}
    \label{tab:occlusion_levels}
    \begin{tabular}{ccc}
    \toprule
    \textbf{Level} & \textbf{Occlusion Ratio Range} & \textbf{Proportion (\%)} \\
    \midrule
    L1  & 0\% -- 10\%     & 12.3\% \\
    L2  & 10\% -- 20\%    & 28.5\% \\
    L3  & 20\% -- 30\%    & 31.6\% \\
    L4  & 30\% -- 40\%    & 19.4\% \\
    L5  & $\geq 40\%$     & 8.2\%  \\
    \bottomrule
    \end{tabular}
\end{table}

Each object in the benchmark is accompanied by its occlusion-level label, enabling fine-grained analysis of model robustness across different occlusion scenarios.


The benchmark and evaluation protocols are designed to provide a standardized and comprehensive platform for occlusion-aware 3D generation methods.

%% file: sec/4_exp.tex
\section{Experiments}

\begin{figure*}[h]
    \centering
    \includegraphics[width=\linewidth]{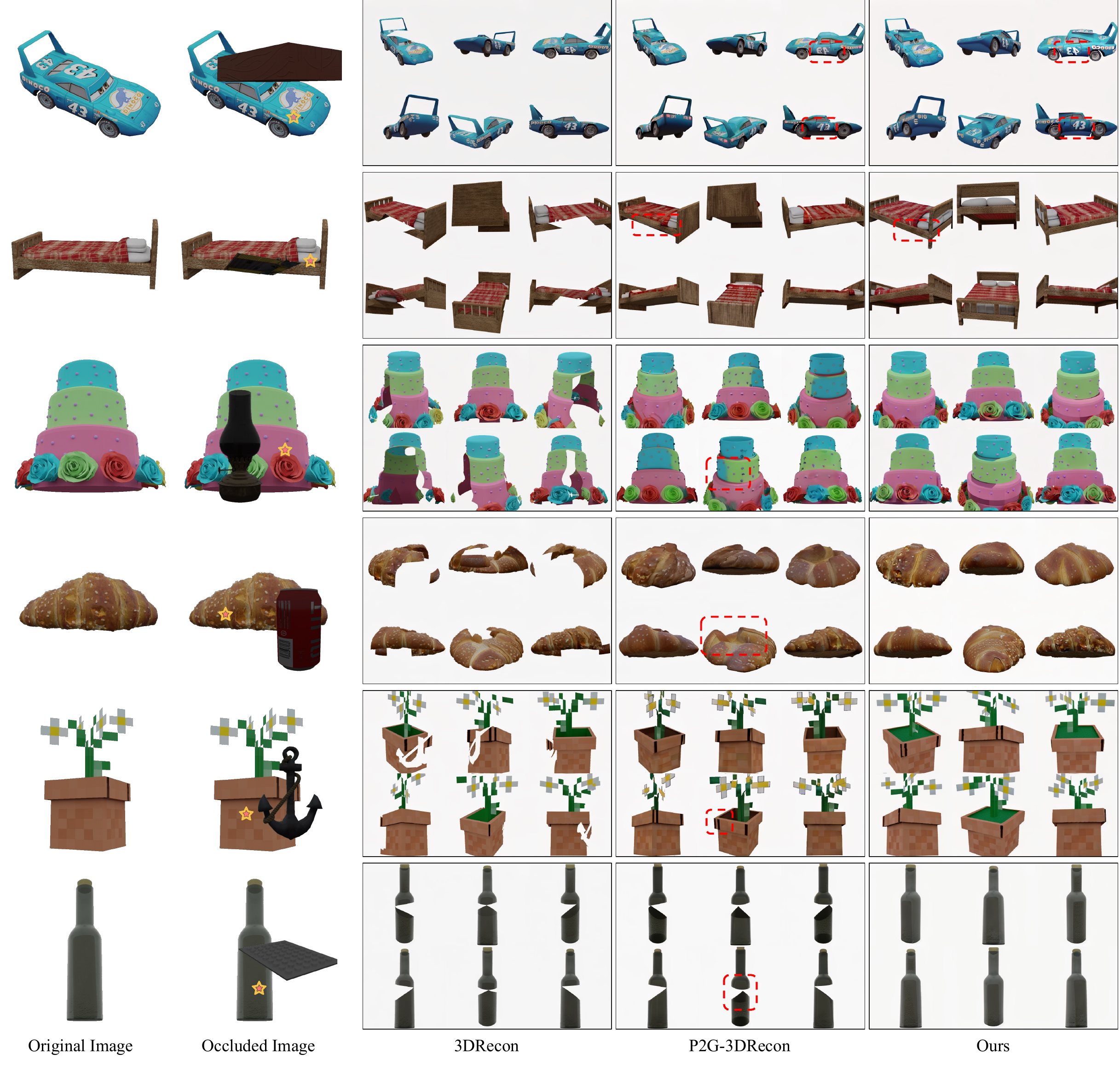}
    \caption{
        Qualitative comparison of de-occlusion results on the Occ-LVIS benchmark. 
        We compare 3DRecon~\cite{shi2023zero123pp,xu2024instantmesh}, P2G-3DRecon~\cite{ozguroglu2024pix2gestalt,shi2023zero123pp,xu2024instantmesh}, and our method. 3DRecon fails to recover occluded content, while P2G-3DRecon suffers from texture degradation and occasional failures. Our method produces consistent, high-quality de-occluded views, enabling high-quality 3D de-occlusion.
}	
    \label{fig:comparison}
\end{figure*}

\subsection{Evaluation Setup}
\textbf{Datasets.}
We conduct quantitative experiments on the proposed Occ-LVIS benchmark (Sec.~\ref{subsec:occ-lvis-benchmark}), 
and perform qualitative evaluations using both Internet images and synthetic generated images. 

\textbf{Baselines.}
We compare our method against the following baselines:
(1) Image-to-3D pipeline (3DRecon): Vanilla Zero123++~\cite{shi2023zero123pp} followed by InstantMesh~\cite{xu2024instantmesh}, without any occlusion handling. 
(2) 2D de-occlusion method + Image-to-3D pipeline (P2G-3DRecon): Pix2Gestalt~\cite{ozguroglu2024pix2gestalt} for 2D amodal completion, followed by Zero123++ and InstantMesh. 
(3) 3D de-occlusion pipeline (Ours): Our proposed method, which jointly performs occlusion-aware view synthesis and downstream 3D reconstruction.

\textbf{Evaluation Metrics.}
We assess the quality of the generated results from both 2D and 3D perspectives.

2D Evaluation: We evaluate the visual fidelity of the generated multi-view images using:
\begin{itemize}
    \item Fréchet Inception Distance (FID)~\cite{heusel2017fid}: Quantifies the distance between the feature distributions of real and synthesized images. Lower values indicate closer alignment in distributional statistics.
    \item Kernel Inception Distance (KID)~\cite{binkowski2018demystifying}: Measures the squared Maximum Mean Discrepancy between real and generated image features with a polynomial kernel. 
    \item CLIP Score~\cite{radford2021clip}: Evaluates the semantic consistency between real and generated images by computing the cosine similarity from a pretrained CLIP model.
\end{itemize}

3D Evaluation: To assess geometric reconstruction quality, we align the generated mesh and ground truth within a unit sphere centered at the origin under consistent coordinate systems. We report:
\begin{itemize}
    \item Chamfer Distance (CD): Computes the average bidirectional distance between points uniformly sampled on the surfaces of the predicted and reference meshes. It reflects surface-level reconstruction fidelity, with lower values indicating better correspondence.
    \item F-Score: Defined as the harmonic mean of precision and recall under a fixed distance threshold, this metric evaluates both completeness and accuracy of the reconstructed geometry. In our evaluation, we report the F1-Score.
    \item Volume Intersection-over-Union (V-IoU): Measures the volumetric overlap between the predicted and ground-truth shapes.
\end{itemize}

\subsection{Quantitative Results}

We report both 2D image quality metrics and 3D geometric fidelity metrics on the Occ-LVIS benchmark 
to evaluate the effectiveness of our method under occluded input settings.

\begin{table}[h]
    \centering
        \caption{2D comparison on the Occ-LVIS benchmark.}

        \begin{tabular*}{0.9\linewidth}{@{\extracolsep{\fill}}cccc}
        \toprule
        Method & CLIP $\uparrow$ & FID $\downarrow$ & KID $\downarrow$ \\
        \midrule
        3DRecon & 0.7430 & 41.3836 & 0.01361 \\
        P2G-3DRecon & 0.7833 & 30.1892 & 0.0043 \\
        Ours & \textbf{0.7892} & \textbf{29.0836} &  \textbf{0.0035} \\
        \bottomrule
    \end{tabular*} 
    \label{tab:2d_comparison}
\end{table}

\begin{table}[h]
    \centering
        \caption{3D comparison on the Occ-LVIS benchmark.}

        \begin{tabular*}{0.9\linewidth}{@{\extracolsep{\fill}}cccc}
        \toprule
        Method & CD $\downarrow$ & F-Score $\uparrow$ & V-IoU $\uparrow$ \\
        \midrule
        3DRecon & 0.0125 & 0.3721 & 0.1565 \\
        P2G-3DRecon & 0.0106 & 0.4470 & 0.3232 \\
        Ours &  \textbf{0.0086} &  \textbf{0.4835} &  \textbf{0.3445} \\
        \bottomrule
    \end{tabular*} 
    \label{tab:3d_comparison}
\end{table}

\textbf{2D Results.} As shown in Table~\ref{tab:2d_comparison}, our method achieves the best performance 
across all metrics. Specifically, it obtains the lowest FID and KID, indicating 
superior perceptual quality and distribution alignment with real images. Additionally, our method 
achieves the highest CLIP similarity score, demonstrating stronger semantic alignment 
between the generated views and the input. Compared to the baseline pipeline (\textit{3DRecon}) and the 
two-stage completion pipeline (\textit{P2G-3DRecon}), our occlusion-aware generation framework 
consistently improves multi-view quality and coherence.

\textbf{3D Results.} We also evaluate the fidelity of reconstructed 3D meshes using Chamfer 
Distance (CD), F-Score, and V-IoU, as reported in Table~\ref{tab:3d_comparison}. Our method 
achieves the lowest CD, indicating accurate surface reconstruction, and the highest F-Score and V-IoU, 
reflecting better geometric completeness and volumetric 
consistency. These results confirm that our multi-view generation preserves structural integrity 
across views and provides high-quality supervision for downstream 3D reconstruction.

\begin{table}[h]
    \centering
    \small
    \caption{Efficiency comparison of different pipelines. Reported metrics indicate resource requirements for a single inference.
}
    \label{tab:efficiency_comparison}
    \begin{tabular*}{0.95\linewidth}{@{\extracolsep{\fill}}cccc}
        \toprule
        Method & Time (s) $\downarrow$ & FLOPs (T) $\downarrow$ & Params (G) $\downarrow$ \\
        \midrule
        3DRecon & 11.09 & 325 & 1.5 \\
        P2G-3DRecon & 20.38 & 332 & 2.7 \\
        Ours & \textbf{11.09} & \textbf{325} & \textbf{1.5} \\
        \bottomrule
    \end{tabular*}
\end{table}

\begin{figure}[h]
    \centering
    \includegraphics[width=\linewidth]{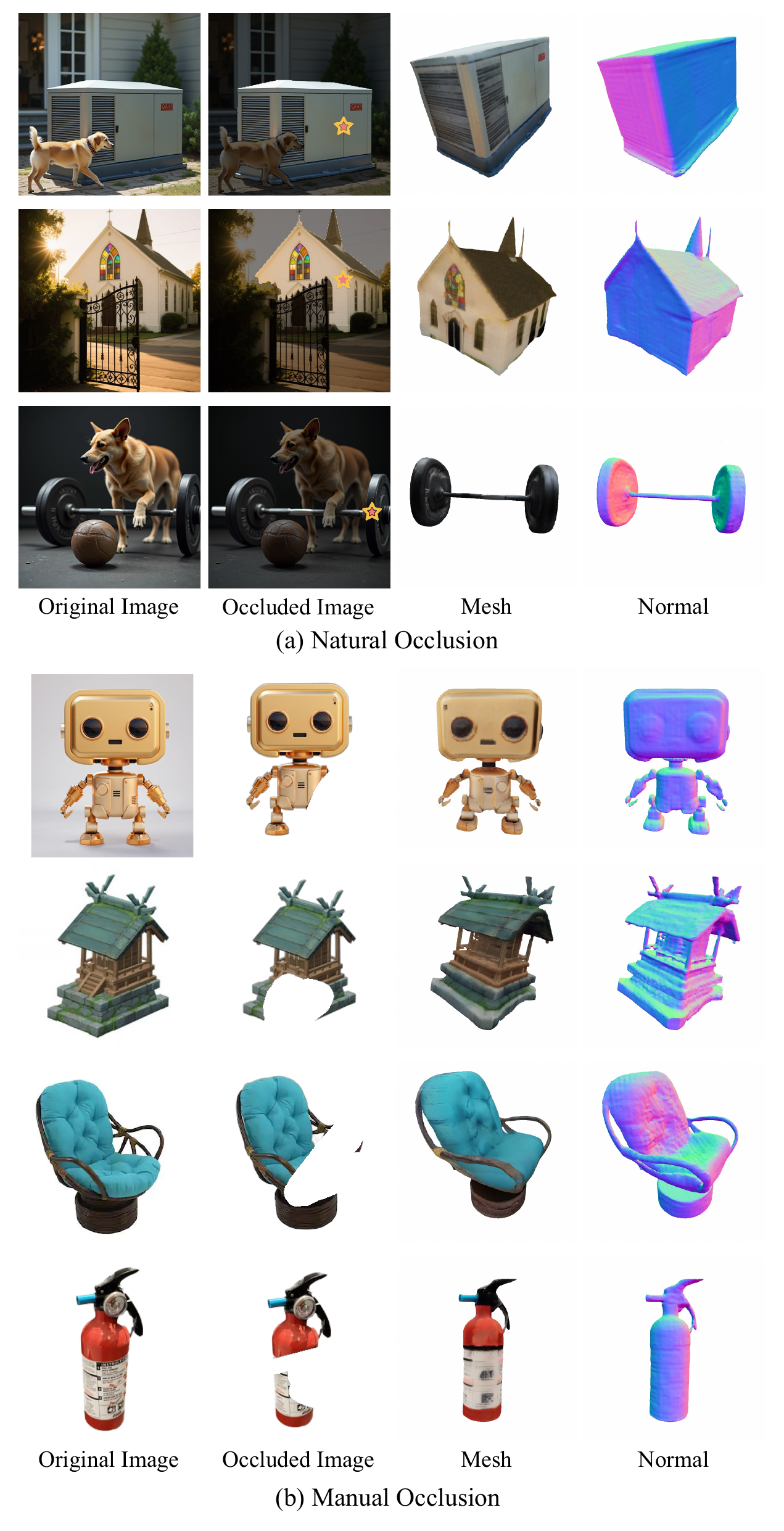}
    \caption{
Qualitative 3D reconstruction results using predicted multi-view de-occlusion images under (a) natural and (b) manual occlusions. 
For each object, we show (left) the original image, (second) the occluded input, 
(third) the reconstructed mesh, and (right) the corresponding surface normals. 
Our method successfully recovers complete geometry and accurate normals despite both real-world and synthetic occlusions, 
demonstrating that our model achieves multi-view consistent occlusion completion.
}
    \label{fig:mv-to-mesh}
    \vspace{-1em}
\end{figure}

\textbf{Efficiency Results.} 
Our method achieves multi-view consistent de-occlusion and 3D reconstruction from a single occluded image in just 11 seconds, 
as shown in Table~\ref{tab:efficiency_comparison}. It maintains efficiency comparable to the \textit{3DRecon} 
pipeline while significantly reducing time cost, FLOPs, and parameter count compared to the \textit{P2G-3DRecon} 
pipeline, providing an efficient and practical solution for real-world 3D de-occlusion applications.

\subsection{Qualitative Results}

As shown in Figure~\ref{fig:results}, we present qualitative results demonstrating the effectiveness of our 
occlusion-aware multi-view generation framework across a wide range of object categories and occlusion scenarios. 
The experiments are conducted on unseen test samples, including both synthetic occlusions and real-world 
occluded photographs. 
Notably, several cases involve severe occlusion exceeding 40\% of the object area, as well as challenging 
visual structures—for instance, the third column shows a “teddy bear holding a book,” and the fourth displays 
an “art installation stacked with firewood.”
Despite these complexities, our method successfully recovers coherent object geometry and 
appearance across all six views. 
The synthesized outputs exhibit strong multi-view consistency, minimal artifacts, and plausible completions 
of the occluded regions. 
These results highlight the robustness and generalization ability of our fine-tuned model, even under 
high occlusion and out-of-distribution conditions.

\textbf{Qualitative Comparison.}
As illustrated in Figure~\ref{fig:comparison}, we present qualitative comparison results 
on the proposed Occ-LVIS benchmark.
The \textit{3DRecon} baseline fails to recover missing content due to the lack of an occlusion completion 
mechanism, resulting in broken or incomplete 3D reconstructions.
While the \textit{P2G-3DRecon} pipeline can complete most occluded regions, it suffers from error 
accumulation across stages. This leads to noticeable texture degradation (e.g., distorted car paint 
in the first row and messy cake colors in the third row), as well as occasional failure cases 
(e.g., missing water bottle content in the sixth row).
In contrast, our method consistently generates high-quality, structure-consistent de-occluded views 
across diverse occlusion types and object categories, enabling more faithful and complete 3D reconstruction.

To further validate the quality and multi-view consistency of our de-occluded image generation, 
we leverage InstantMesh~\cite{xu2024instantmesh} to reconstruct 3D geometry from the predicted 
six-view outputs. We also visualize the surface normals of the resulting mesh for qualitative inspection. 
As shown in Figure~\ref{fig:mv-to-mesh}, our method enables high-quality 3D reconstruction across 
various scenarios, producing geometrically consistent multi-view completions and realistic object surfaces. 
These results confirm that our generated views not only complete occluded content plausibly, 
but also serve as reliable inputs for downstream mesh-based 3D modeling.

%% file: sec/5_conclusion.tex
\section{Limitations and Conclusion}

\textbf{Limitations.}  
While our method demonstrates strong performance across diverse occlusion scenarios, it still struggles under extremely heavy occlusions where structural cues are severely missing. In addition, although we use category-agnostic training data, the model may still perform suboptimally on unseen object categories with highly unusual shapes or topologies.

\textbf{Conclusion.}  
We present DeOcc-1-to-3, a self-supervised paradigm for multi-view 3D de-occlusion from a single occluded image. 
By fine-tuning a multi-view diffusion model, our method jointly completes missing structures and synthesizes 
geometrically consistent novel views. It requires no architectural modifications, integrates seamlessly with existing 
3D reconstruction pipelines, and generalizes well to a variety of occlusion scenarios. 
To support future research, we also introduce a benchmark for standardized evaluation of occlusion-aware 3D modeling.